%% file: Paper.tex
\documentclass[a4paper, 10pt, conference]{ieeeconf}      %
\usepackage{FG2025}

\FGfinalcopy %

\usepackage{caption}
\usepackage{graphics} %
\usepackage{graphicx}
\usepackage{mathptmx} %
\usepackage{amsmath} %
\usepackage{amssymb}  %
\usepackage{algorithm}
\usepackage{algpseudocode}
\usepackage{booktabs}
\usepackage{xcolor}  
\usepackage{colortbl}  
\usepackage{multirow} 
\usepackage{eucal}

\def\FGPaperID{101} %

\title{\LARGE \bf
EmoGene: Audio-Driven Emotional 3D Talking-Head Generation 
}

\author{\parbox{16cm}{\centering
    {\large Wenqing Wang$^1$ and Yun Fu$^{1,2}$}\\
    {\normalsize
    $^1$ Khoury College of Computer Science, Northeastern University, USA\\
    $^2$ Department of Electrical and Computer Engineering, Northeastern University, USA}}
}

\begin{document}

\ifFGfinal
\thispagestyle{empty}
\pagestyle{empty}
\else
\author{Anonymous FG2025 submission\\ Paper ID \FGPaperID \\}
\pagestyle{plain}
\fi
\maketitle

\begin{abstract}
Audio-driven talking-head generation is a crucial and useful technology for virtual human interaction and film-making. While recent advances have focused on improving image fidelity and lip synchronization, generating accurate emotional expressions remains underexplored. In this paper, we introduce EmoGene, a novel framework for synthesizing high-fidelity, audio-driven video portraits with accurate emotional expressions. Our approach employs a variational autoencoder (VAE)-based audio-to-motion module to generate facial landmarks, which are concatenated with emotional embedding in a motion-to-emotion module to produce emotional landmarks. These landmarks drive a Neural Radiance Fields (NeRF)-based emotion-to-video module to render realistic emotional talking-head videos. Additionally, we propose a pose sampling method to generate natural idle-state (non-speaking) videos for silent audio inputs. Extensive experiments demonstrate that EmoGene outperforms previous methods in generating high-fidelity emotional talking-head videos.
\end{abstract}

\section{INTRODUCTION}

Generating audio-driven talking-head videos is an important technology across a variety of applications, including digital humans, virtual reality, and entertainment. Achieving realistic talking-head synthesis requires not only accurate lip synchronization and high-fidelity visuals but also accurate emotional expressions. Over the past few years, numerous methods \cite{depthaware, facial, livespeechportraits, videoretalking, makeittalk, oneshot, wav2lip} have been proposed to generate audio-driven talking-head videos. However, they often suffer from artifacts, unrealistic visuals, or failure to capture the details of the target identity. For instance, Wav2Lip \cite{wav2lip} demonstrates robust lip synchronization, but it is unable to generate full facial regions, limiting its generalization. DaGAN \cite{depthaware} adopts a depth-conditioned generative adversarial network (GAN) to generate talking heads, but its GAN-based renderer struggles with training instability and modeling intricate details. FACIAL \cite{facial} integrates facial implicit attribute learning with GANs but encounters similar issues with training stability and detail generation.

Recent advancements \cite{nerf2024_1, nerf2024_2, syntalk, ernerf, adnerf, geneface} in Neural Radiance Fields (NeRF) \cite{nerf} have significantly improved talking-head generation by enabling high-fidelity video synthesis with precise details. Notable works such as ER-NeRF \cite{ernerf}, GeneFace++ \cite{geneface++}, and SyncTalk \cite{syntalk} enhance lip synchronization while maintaining high-quality rendering. Due to its efficient rendering speed, 3D Gaussian Splatting \cite{3dgs} has been recently adopted in synthesizing talking-head videos \cite{gstalker, pointtalk, gaussiantalker, talkinggaussian}. For instance, works such as GaussianTalker \cite{gaussiantalker} and TalkingGaussian \cite{talkinggaussian} leverage the deformation of 3D Gaussian attributes to generate talking heads efficiently. Despite these advancements, the emotional aspect of generating vivid talking-head videos has been overlooked.

To address this gap, recent studies \cite{eat, eamm, emo_2024_2, evp, emo_2024_3, emo_2024_4, emo_2024_5, emotalk, emoface} have begun focusing on emotion-aware talking-head synthesis. For example, EAT \cite{eat} uses an emotion deformation model for controlled emotional video generation, while FlowVQTalker \cite{emo_2024_3} employs normalizing flows and vector quantization to synthesize emotional talking heads. EAMM \cite{eamm} guides emotional talking-head generation using emotional and pose videos, and FG-EmoTalk \cite{emo_2024_5} utilizes disentangled expression latent codes and facial features. EVP \cite{evp} disentangles speech content and emotional features for emotional video portraits, and PC-AVS \cite{pc_avs} incorporates pose control.

While these works made improvements in generating emotional expressions, they struggle with several challenges in terms of generating vivid emotions. 1) \textbf{Emotional accuracy}. Although previous methods can generate some forms of emotional expressions, their generated emotional expressions are often partially inaccurate or invisible. For instance, some approaches \cite{eat} generate correct brow movements but mismatched mouth expressions, reducing the realism of synthesized videos. 2) \textbf{Identity preservation}. Many methods fail to preserve the target identity during emotional representation learning, leading to facial distortions and diminished video fidelity. 3) \textbf{Idle state}. Given silent audio, existing works often generate extra lip movements or unnatural body motions, either introducing excessive motion or remaining entirely static. This constraint limits their capability to produce natural idle-state videos in many applications, such as digital humans and conversational virtual assistants.

To address these challenges, we introduce \textit{EmoGene}, a novel pipeline for generating emotional talking-head videos driven by audio inputs. For emotional accuracy, we propose a motion-to-emotion module. This module is composed of a landmark deformation model (LDM), which is a neural network trained on an emotion-labeled video dataset MEAD \cite{mead} to generate accurate emotional facial landmarks from emotion text labels and neutral landmarks. For identity preservation, we utilize an emotion-to-video module that consists of NeRF models to preserve the target person's identity while rendering high-fidelity videos conditioned on emotional landmarks. To address the idle-state challenge, we develop a pose sampling method that produces natural idle-state videos from silent audio inputs. This can enhance the realism and naturalness of the synthesized talking-head videos.

The main contributions of our work are as follows:

\begin{itemize}
  \item We present a three-stage framework for generating audio-driven emotional talking-head videos with accurate emotions and preserved identity.

  \item We introduce a landmark deformation model for controllable emotional landmark generation.

  \item We propose a pose sampling method that produces natural idle-state videos.

  \item Experiments show that EmoGene outperforms previous methods in generating emotional, high-fidelity talking-head videos.
\end{itemize}

\section{RELATED WORK}

\subsection{Audio-Driven Talking-Head Generation} 

Generating audio-driven talking heads has attracted considerable interest recently \cite{wav2lip, nerf2024_2, depthaware, facial, livespeechportraits, makeittalk, oneshot, videoretalking}. Early approaches \cite{lip_1, lip_2, lip_3, wav2lip, talklip} primarily focus on synthesizing lip motion. For example, Wav2Lip \cite{wav2lip} employs a lip-sync model to achieve accurate lip motions but concentrates solely on the lip region, resulting in reduced facial coherence and limited generalization. To generate more realistic talking heads, full-face synthesis methods \cite{tk_1, flowguided, videoretalking, facial} have been introduced. Nonetheless, purely audio-driven techniques often lack explicit 3D geometric guidance for pose control, which can compromise the robustness and stability of generated videos. For instance, Zhang \emph{et al.} \cite{flowguided} proposes an audio-driven talking-head framework with guided flow, but inaccurate foreground masks can degrade its generation quality. Similarly, LiveSpeechPortrait \cite{livespeechportraits} renders talking-head videos from audio and motion features through an image translation framework, but it struggles with artifacts and limited generalization.

To improve generation robustness and stability, several works \cite{nocentini2023learninglandmarksmotionspeech, a2l_1, a2l_2, makeittalk, a2l_3, a2l_4} have explored converting audio into facial landmarks as intermediate representations for driving talking-head synthesis. Berretti \emph{et al.} \cite{nocentini2023learninglandmarksmotionspeech} utilizes facial landmark motion derived from audio to synthesize 3D talking heads. DreamHead \cite{a2l_1} predicts facial landmarks from audio to guide a diffusion model for portrait video generation. Many approaches adopt 3D Morphable Models (3DMMs) \cite{3dmm} as reliable and accurate landmark representations. Inspired by these works, our framework leverages audio-driven 3DMM landmarks as intermediate representations to enhance expression controllability and geometric accuracy in talking-head generation.

Since the introduction of NeRF \cite{nerf}, NeRF-based methods \cite{nerf2024_1, nerf2024_2, syntalk, ernerf, adnerf, geneface, geneface++} have advanced talking-head rendering due to their high-quality results. ER-NeRF \cite{ernerf} uses tri-plane hash encoders to render talking heads conditioned on audio features. Synctalk \cite{syntalk} builds on tri-plane NeRF with a Face-Sync Controller and facial blendshape model to improve generation synchronization. GeneFace++ \cite{geneface++} integrates grid-based NeRF with audio-driven motion for realistic rendering and robust lip synchronization. Building on these works, our EmoGene framework adopts NeRF for high-quality, realistic talking-head generation. To reduce rendering complexity, some recent works \cite{gstalker, pointtalk, gaussiantalker, talkinggaussian} have leveraged 3D Gaussian Splatting to generate talking heads. Methods such as GaussianTalker \cite{gaussiantalker} and TalkingGaussian \cite{talkinggaussian} deform attributes of 3D Gaussian primitives to generate talking heads. However, the point-based nature of Gaussian Splatting often leads to artifacts and unstable rendering results. Despite these advancements, emotion generation in talking-head synthesis remains underexplored.

\subsection{Emotion-Aware Talking-Head Generation}

\begin{figure*}[!t]
  \centering
  \includegraphics[width=\textwidth, trim=0 0 0 0, clip]{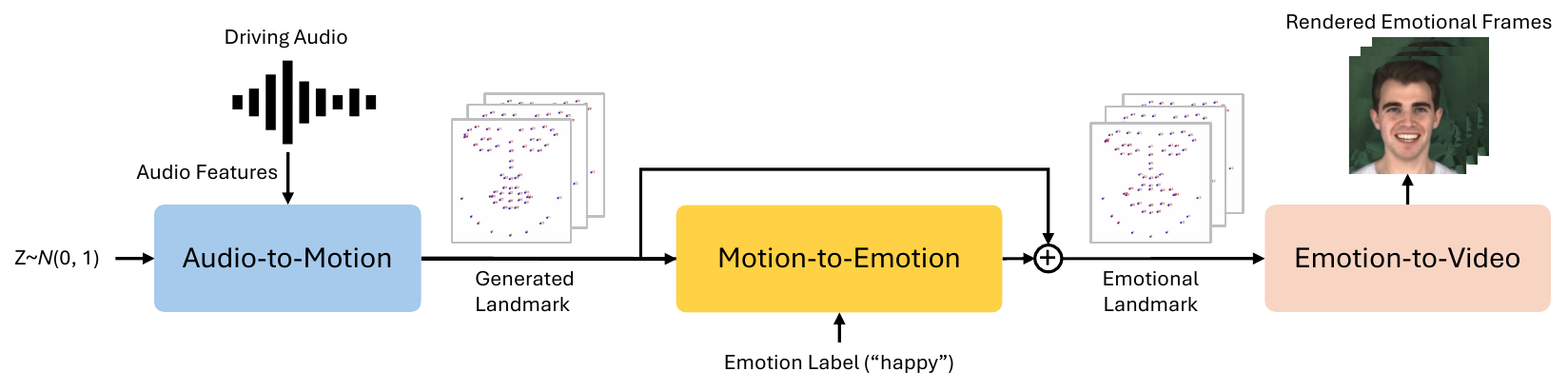}
  \caption{\textbf{EmoGene pipeline.} 1) The Audio-to-Motion module converts input audio features into neutral facial landmarks. 2) The Motion-to-Emotion module transforms these landmarks into emotional landmarks based on the emotion label. 3) The Emotion-to-Video module generates the emotional talking-head video conditioned on the emotional landmarks.}
  \label{Pipeline}
\end{figure*}

Emotion-aware talking-head generation extends audio-driven talking-head methods by incorporating emotional features to produce lifelike emotional expressions. Unlike traditional approaches that prioritize audio-lip synchronization, this task requires that the generated talking heads not only have accurate audio-lip alignment but also vividly display specific emotional expressions. The goal is to produce videos that are both visually convincing and emotionally authentic, maintaining the identity and natural appearance of the subject while dynamically reflecting the intended emotions.

Previous works \cite{eat, eamm, emo_2024_3, emo_2024_4, emo_2024_5, styletalk, evp, emotalk, emoface} have explored diverse methodologies to achieve controllable emotional expressions in talking-head generation. These approaches typically leverage deep learning techniques to model and manipulate emotional features. For instance, EAT \cite{eat} employs a transformer-based emotion deformation network that uses emotional features and source latent tokens to guide the generation of emotionally expressive talking heads. Other methods, such as EAMM \cite{eamm} and StyleTalk \cite{styletalk}, may incorporate style-based GANs or disentangled feature representations to separate and control emotional attributes from identity and audio-driven motion.

Despite these advancements, existing methods face significant challenges in achieving a balance between emotional accuracy and high image fidelity. Emotional accuracy refers to the precise and consistent portrayal of the intended emotion across all facial regions, ensuring that expressions appear natural and coherent. High image fidelity, on the other hand, relates to the visual quality of the generated video. For example, while EAT \cite{eat} successfully generates emotionally expressive talking heads, it often struggles with issues such as identity distortion, where the subject’s facial features deviate from the original, and inconsistent emotional expressions, where different parts of the face (e.g., brows, eyes, mouth) fail to convey the same emotion cohesively. Similarly, other methods like \cite{emo_2024_3, emo_2024_4} may produce high-quality visuals but fall short in capturing smooth emotional transitions, leading to unnatural or exaggerated expressions.

These limitations highlight the need for emotion-aware talking-head generation methods that can simultaneously achieve emotional accuracy and high image fidelity. This paper aims to address these challenges by introducing a novel framework that balances these objectives, providing an effective method for creating lifelike and emotionally compelling talking-head videos. By addressing the shortcomings of prior works, this research seeks to establish a strong foundation for future advancements in realistic emotional talking-head generation, with diverse applications in digital avatars, human-computer interaction, entertainment, and beyond.

\section{METHOD}

In this section, we present our EmoGene framework. As illustrated in Figure~\ref{Pipeline}, EmoGene consists of three components: 1) Audio-to-Motion: a variational autoencoder (VAE) that converts audio features into neutral facial landmarks; 2) Motion-to-Emotion: a landmark deformation model that transforms neutral landmarks into emotional landmarks; 3) Emotion-to-Video: NeRF models that render the emotional talking-head video.

\begin{figure}[!t]
  \centering
  \includegraphics[width=\columnwidth]{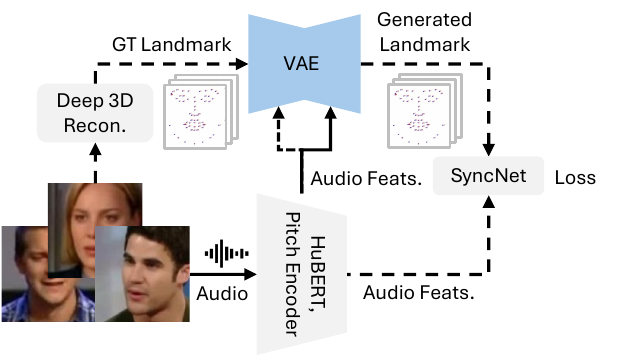}
  \caption{\textbf{The overview of the audio-to-motion module.} The dashed arrow indicates that the process occurs only during training.}
  \label{a2m}
\end{figure}

\subsection{Audio-to-Motion}
We utilize a VAE model \cite{vae, geneface, geneface++} to generate audio-driven facial landmarks, as presented in Figure~\ref{a2m}. The VAE model is conditioned on ground-truth (GT) facial landmarks and audio features, with a pre-trained SyncNet \cite{syncnet} ensuring accurate alignment between audio and landmarks.

\textbf{Encoder and Decoder.} The encoder and decoder are structured as convolutional neural networks based on a WaveNet architecture \cite{wavenet}. Progressively increasing convolutional layers expand the receptive field, enabling efficient generation of variable-length landmark sequences.

\textbf{Training Process.} During training, GT 3D facial landmarks are extracted from videos using Deep 3D Face Reconstruction \cite{deep3drecon}, from which 68 key landmark points are selected to represent facial motion. Audio features, including HuBERT \cite{hubert} and pitch, are also extracted. The VAE is trained using a Monte-Carlo ELBO loss \cite{detectoutofdistri} for training efficiency, guided by a pre-trained SyncNet \cite{syncnet} to enforce audio-lip synchronization of the generated landmarks. The training loss for the VAE is defined as:
\begin{equation}
\mathcal{L}_{\text{VAE}} = \mathbb{E}\left[\|l - \hat{l}\|_2^2 + \text{KL}(z \mid \hat{z}) + \mathcal{L}_{\text{Sync}}(a, \hat{l})\right].
\end{equation}

In this setup, $l$ represents the GT facial landmark, $\hat{z}=\text{Encoder}(l, a)$ denotes the latent encoding, $a$ represents the input audio, $z \sim N(0,1)$ is the sampled latent code, and $\hat{l}=\text{Decoder}(\hat{z}, a)$ denotes the generated landmark. $\text{KL}$ is the Kullback-Leibler divergence, and $\mathcal{L}_{Sync}$ denotes the lip-synchronization loss computed by SyncNet. During inference, only the decoder is required to generate audio-driven facial landmarks.

\subsection{Motion-to-Emotion}
\begin{figure}[!t]
  \centering
  \includegraphics[width=\columnwidth]{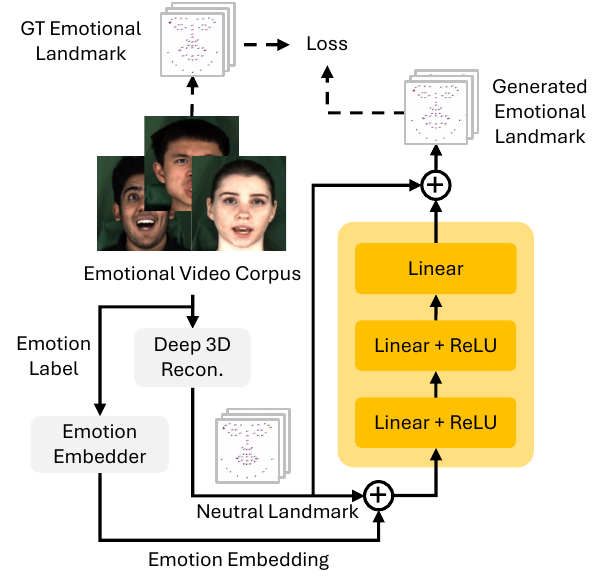}
  \caption{\textbf{The overview of motion-to-emotion module.} The dashed arrow indicates that the process occurs only during training.}
  \label{m2e}
\end{figure}

Because the audio-to-motion training process is not conditioned on emotion, it can only generate neutral landmarks. To enable the generation of emotionally expressive landmarks, we introduce the motion-to-emotion module (Figure~\ref{m2e}). This module transforms neutral landmarks into emotional landmarks using a landmark deformation model, which generates the landmark deformation displacements for obtaining the emotional landmarks.

\begin{figure*}[!t]
  \centering
  \includegraphics[width=\textwidth, trim=0 0 0 0, clip]{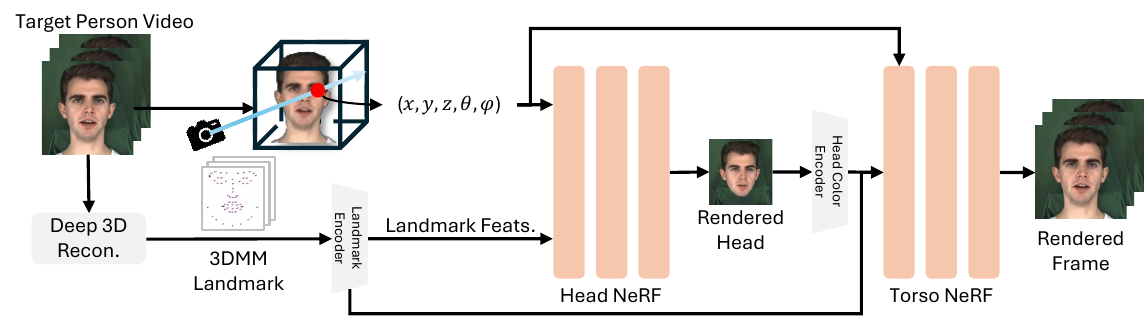}
  \caption{\textbf{The overview of the emotion-to-video module.} The NeRF models are trained to render the talking-head video from the driving landmarks.}
  \label{e2v}
\end{figure*}

\textbf{Landmark Deformation Model.} To generate emotional expressions, we design a neural network that consists of 3 fully connected (FC) layers. The landmark deformation model (LDM) takes concatenated neutral landmarks and emotional embeddings as input and outputs landmark deformation displacements. The LDM is defined as:
\begin{equation}
LDM(x) = FC(ReLU(FC(ReLU(FC(x))))),
\end{equation}
where $x$ denotes the concatenated input of neutral landmark and emotional embedding. The emotional landmark is synthesized by concatenating the neutral landmark with the landmark deformation displacement:
\begin{equation}
\hat{l}_E = \hat{l} \oplus \Delta \hat{l},
\end{equation}
where $\hat{l}_E$ is the generated emotional landmark, $\hat{l}$ is the neutral landmark, and $\Delta \hat{l}$ is the landmark deformation displacement.

\textbf{Training Process.} During training, GT emotional landmarks are extracted from video data, and an emotion embedder generates emotion embeddings from corresponding emotion labels. The neutral landmarks and emotional embeddings are concatenated and fed into the landmark deformation model to produce deformation displacements. These displacements are concatenated with the neutral landmarks to generate emotional landmarks. The training loss for the LDM is:
\begin{equation}
\mathcal{L}_{\text{LDM}} = \mathbb{E}\left[\|l_E - \hat{l}_E\|_2^2\right],
\end{equation}
where $l_{E}$ is the GT emotional landmark, and $\hat{l}_E$ is the generated emotional landmark.

\subsection{Emotion-to-Video}
\begin{figure*}[!t]
  \centering
  \includegraphics[width=\textwidth, trim=0 0 0 0, clip]{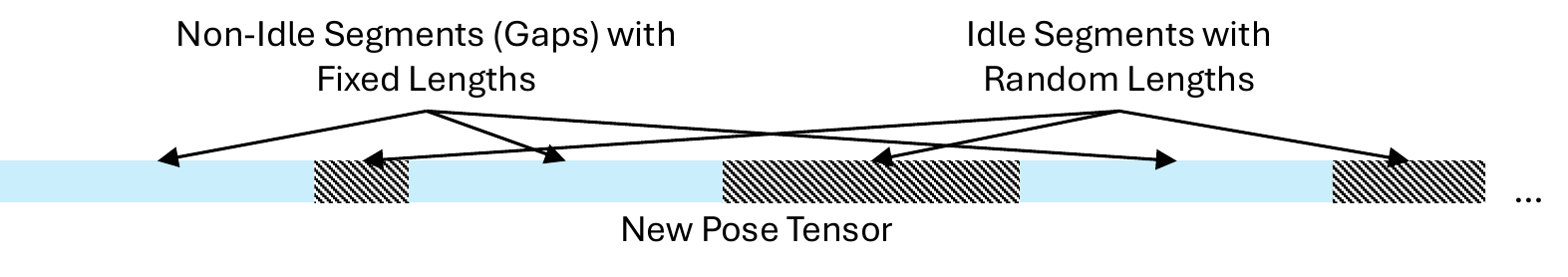}
  \caption{\textbf{New pose tensor construction.} To construct the new pose tensor, the idle pose segments are inserted after their corresponding non-idle pose tensors.}
  \label{idle1}
\end{figure*}

\begin{figure}[!t]
  \centering
  \includegraphics[scale=0.55]{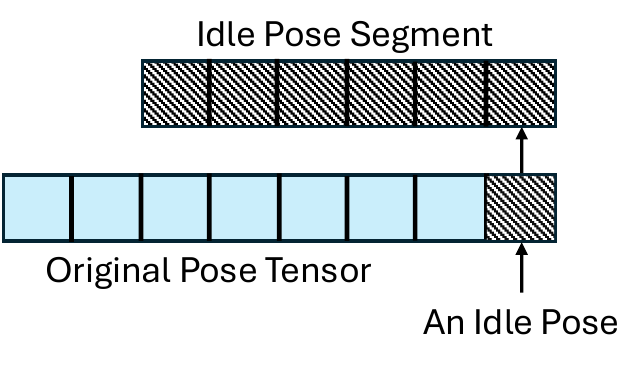}
  \caption{\textbf{Idle pose segment construction.} Idle pose segments are created by replicating the identified idle poses from the original pose tensor.}
  \label{idle2}
\end{figure}

The emotion-to-video module (Figure~\ref{e2v}) leverages NeRF-based models \cite{nerfo5, geneface, geneface++} to synthesize the emotional talking-head video from the generated emotional landmarks.

\textbf{Emotion Landmark-Conditioned Head-NeRF.} To dynamically render high-quality talking-head videos, we leverage a NeRF-based model \cite{nerfo5, geneface++}, which extends the original NeRF \cite{nerf} by conditioning on the 3D location $x$, viewing direction $d$, and emotional landmark $\hat{l}_E$:
\begin{equation}
F : (x, d, \hat{l}_E) \to (c, \sigma),
\end{equation}
where $c$ represents the estimated RGB color and $\sigma$ denotes the estimated volume density in the 3D neural radiance field.

To render each pixel of the image frame, we aggregate the color $c$ along the ray $r(t)=o+t \cdot d$, originating from the camera at $o$, using differentiable volume rendering:
\begin{equation}
C(r, \hat{l}_E) = \int_{t_n}^{t_f} \sigma(r(t), \hat{l}_E) \cdot c(r(t), \hat{l}_E, d) \cdot T(t) \, dt,
\end{equation}
where $C$ is the rendered RGB color of the pixel for the emitted ray $r$ in the 3D space. $t_n$ and $t_f$ are the near and far bounds of the ray. $T(t)$ is the accumulated transmittance along the ray from $t_n$ to the current point $t$:
\begin{equation}
T(t) = \exp\left(-\int_{t_n}^t \sigma(r(s), \hat{l}_E) \, ds\right),
\end{equation}
where $s$ is an intermediate point on the ray from $t_n$ to $t$.

\textbf{Torso-NeRF.} To ensure seamless head-torso integration, the torso-NeRF is conditioned on the head-NeRF's output color. This enables the torso-NeRF to receive more information about the rendered head results, which strengthens the rendering coherence of the head and torso and enhances the realism of the generated videos. The torso-NeRF model is defined as:
\begin{equation}
F_{torso} : (x, C_{head}; d_0, P, \hat{l}_E) \to (c, \sigma).
\end{equation}

Here, $d_0$ is the canonical view direction, and $P \in \mathbb{R}^{3 \times 4}$ represents the head pose consists of a rotation matrix combined with a translation vector.

\textbf{Training Process.} To train our NeRF models, we extract the facial landmarks and images from the input video frames. We then utilize the landmark-image pairs to train the NeRF models. The training objective for the NeRF models is to minimize the $L_2$ reconstruction error between the generated and ground-truth images:
\begin{equation}
\mathcal{L}_{\text{NeRF}} = \mathbb{E}\left[\|C(r, \hat{l}_E) - C_{GT}\|_2^2\right].
\end{equation}

\begin{algorithm}
\caption{Generate idle segments with random idle lengths and fixed gap length.}
\label{algo_pose_samp}
\begin{algorithmic}[1]
\Function{Segments}{$n, \text{len}_\text{min}, \text{len}_\text{max}, \text{len}_\text{gap}$}
    \State $\text{segments} \gets []$
    \State $\text{position}_\text{curr} \gets 0$
    \While{$\text{position}_\text{curr} < n$}

        \State $l \gets Random(\text{len}_\text{min},\min(\text{len}_\text{max}, n - \text{position}_\text{curr}))$

        \State $\text{start} \gets \text{position}_\text{curr}, \text{end} \gets \text{start} + l - 1$
        \If{$\text{end} + \text{len}_\text{gap} \geq n$} \textbf{break} \EndIf
        \State $\text{segments}.append(\text{start}, \text{end})$
        \State $\text{position}_\text{curr} \gets \text{end} + 1 + \text{len}_\text{gap}$
    \EndWhile
    \State \Return $\text{segments}$
\EndFunction
\end{algorithmic}
\end{algorithm}

\subsection{Pose Sampling Method for Idle State}
We propose a pose sampling algorithm (Algorithm~\ref{algo_pose_samp}) to synthesize natural body and lip movements for idle-state videos. As shown in Figure~\ref{idle2}, the algorithm begins by identifying initial idle poses within the original pose tensor. These poses are then replicated to form a set of idle pose segments with random idle lengths and fixed gap length (the number of non-idle poses) between them. To create a new pose tensor incorporating the idle state, we insert these generated idle pose segments into the original tensor, as shown in Figure~\ref{idle1}.

\textbf{Determining Minimum and Maximum Idle Segment Lengths.} To generate idle segments, we develop a method to determine the minimum and maximum lengths of idle segments. Given a pose tensor $X = \{x_1, \ldots, x_i, x_{i+1}, \ldots, x_{n-1}\}$, this method computes the cosine similarities between consecutive poses at indices $i$ and $i+1$ as follows:
\begin{equation}
\text{CosineSimilarity}_i = \frac{\mathbf{x}_{i} \cdot \mathbf{x}_{i+1}}{\|\mathbf{x}_{i}\| \|\mathbf{x}_{i+1}\|}.
\end{equation}

A pair of consecutive poses is classified as part of an idle segment if their cosine similarity is equal to 1, indicating that the two poses are aligned and have maximum similarity. By comparing the lengths of these idle segments, the method determines the minimum and maximum idle segment lengths.

\textbf{Generating Idle Segments with Random Lengths and Fixed Gaps.} As shown in Algorithm~\ref{algo_pose_samp}, the algorithm generates pairs of (\textit{start}, \textit{end}) indices corresponding to the idle segments within the pose tensor. To ensure natural idle-state motion, the algorithm iterates through the pose tensor and randomly determines the length of each idle segment within the constraints of the minimum and maximum idle segment lengths. This random sampling introduces variability in the idle motion, enhancing the naturalness of the generated idle-state videos. After each idle segment, the algorithm keeps a fixed gap before the next idle segment begins. This algorithm produces idle motion periods that emulate natural waiting or resting states, which can enhance the realism of idle-state videos.

\textbf{Inserting Idle Segments into the Pose Tensor.} To integrate the idle state, we create a new pose tensor by inserting the generated idle segments after their corresponding non-idle pose tensors (Figure~\ref{idle1}). This process produces natural idle-state motions and smooth motion-stillness transitions for silent audio inputs, resulting in more realistic rendered videos.

\input{tables/quant_main}

\input{tables/quant_idle}

\section{EXPERIMENTS}

\subsection{Experimental Settings}

\textbf{Datasets.} We trained the audio-to-motion module on VoxCeleb2 \cite{voxceleb2}, which contains more than 1 million speech utterances from 6,112 identities, to learn a generalized audio-to-motion mapping. For learning the landmark deformation model, we trained it on the MEAD dataset \cite{mead}, which includes labeled emotional talking videos of 60 identities with 8 emotions (happy, sad, angry, surprise, disgust, neutral, contempt, fear) for each utterance. We divided the MEAD dataset into training and testing sets based on identity. For NeRF model training, we utilized the videos from the MEAD dataset and the out-of-domain (ODD) video dataset \cite{adnerf}. Each training video has a duration of 3–6 minutes, a resolution of 512×512, and a frame rate of 25 FPS.

\textbf{Baselines.} We compare EmoGene with 7 notable works: a transformer-based method, EAT \cite{eat}; a pre-trained lip-synchronization method, Wav2Lip \cite{wav2lip}; three NeRF-based methods (GeneFace++ \cite{geneface++}, SyncTalk \cite{syntalk}, ER-NeRF \cite{ernerf}); two 3D Gaussian Splatting-based methods (GaussianTalker \cite{gaussiantalker} and TalkingGaussian \cite{talkinggaussian}).

\textbf{Implementation Details.} EmoGene was trained on an NVIDIA RTX A6000 GPU. The VAE and the landmark deformation model took approximately 40K iterations to converge (about 14 hours). The NeRF models were trained for 400K iterations (about 12 hours).

\subsection{Quantitative Evaluation}

\textbf{Metrics.} We assess the image quality and fidelity of generated videos using SSIM and PSNR. To evaluate the emotional accuracy, we compute the average predicted probability of the target emotion across all video frames using DeepFace \cite{taigman2014deepface}. We employ the landmark distances for the mouth (M-LMD) and face (F-LMD) to measure audio-lip synchronization and expression accuracy. To evaluate motion naturalness and consistency in idle-state videos, we analyze the average and standard deviation of motion velocity and acceleration between poses. Motion velocity, formulated as $v(t) = x(t+1) - x(t)$, measures the change in pose between consecutive video frames, where $x(t)$ denotes the pose at timestep $t$. Motion acceleration, defined as $a(t) = v(t+1) - v(t)$, measures the change in velocity over time.

\textbf{Evaluation Results.} 
The quantitative results are shown in Table~\ref{table_quant} and Table~\ref{table_quant_idle} respectively. We have the following observations: 1) EmoGene achieves high image quality and fidelity while obtaining high emotion and expression accuracy with effective lip synchronization. As shown in Table~\ref{table_quant}, EmoGene outperforms the other methods on MEAD dataset in PSNR, emotion score, and M/F-LMD. It achieves the third best in SSIM, potentially due to deformations in emotional landmarks. On the OOD dataset, EmoGene achieves the best scores in SSIM and M/F-LMD, the second-best emotion score, and the third-best PSNR; 2) EmoGene generates natural and consistent idle-state motions. In Table~\ref{table_quant_idle}, EmoGene achieves the best scores in average motion velocity, average motion acceleration, and acceleration standard deviation. In addition, it also obtains the second-best velocity standard deviation, likely influenced by the inherent randomness in its idle length sampling process.

\begin{figure*}[!t]
  \centering
  \includegraphics[width=\textwidth, trim=0 0 0 0, clip]{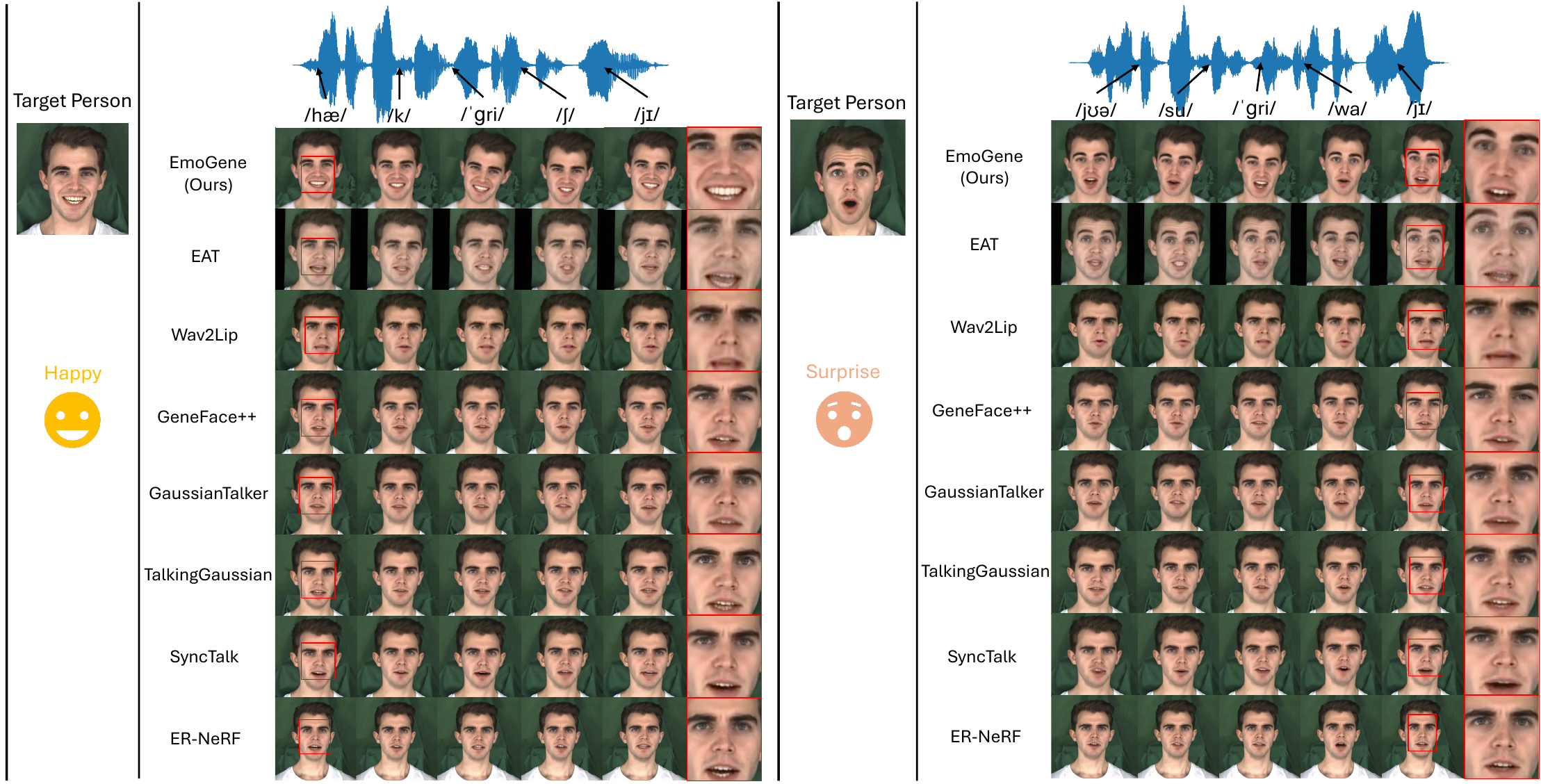}
  \caption{\textbf{Qualitative comparison of generated keyframe results.} Results for \textit{happy} and \textit{surprise} are shown in the left and right columns, respectively. The top row displays the phonetic symbols corresponding to the talking-head results generated by the evaluated methods.}
  \label{hap_sur}
\end{figure*}

\input{tables/mos}

\subsection{Qualitative Evaluation}
To demonstrate the qualitative results of the evaluated methods, we present the keyframes from two emotion-specific clips in Figure~\ref{hap_sur}. We observe that EmoGene demonstrates accurate facial emotional expressions, precise lip synchronization, and strong identity preservation. In contrast, NeRF-based methods (GeneFace++, SyncTalk, and ER-NeRF) and 3D Gaussian Splatting-based methods (GaussianTalker, TalkingGaussian) achieve effective lip-synchronization and facial reconstruction, but they are unable to generate emotions. Wav2Lip obtains robust lip-synchronization, but it only generates the mouth region instead of the entire face, and its generated mouth results are blurry and do not preserve the mouth features of the target person. Although EAT can synthesize localized emotional expressions, it struggles to generate consistent emotional expressions across all facial regions (e.g., brows, eyes, and mouth), which leads to unnatural expressions. Furthermore, the generated images of EAT exhibit facial feature distortions, resulting in identity loss and reduced image fidelity, with noticeable deviations in facial structure compared to the target person.

\begin{figure}[!t]
  \centering
  \includegraphics[width=\columnwidth]{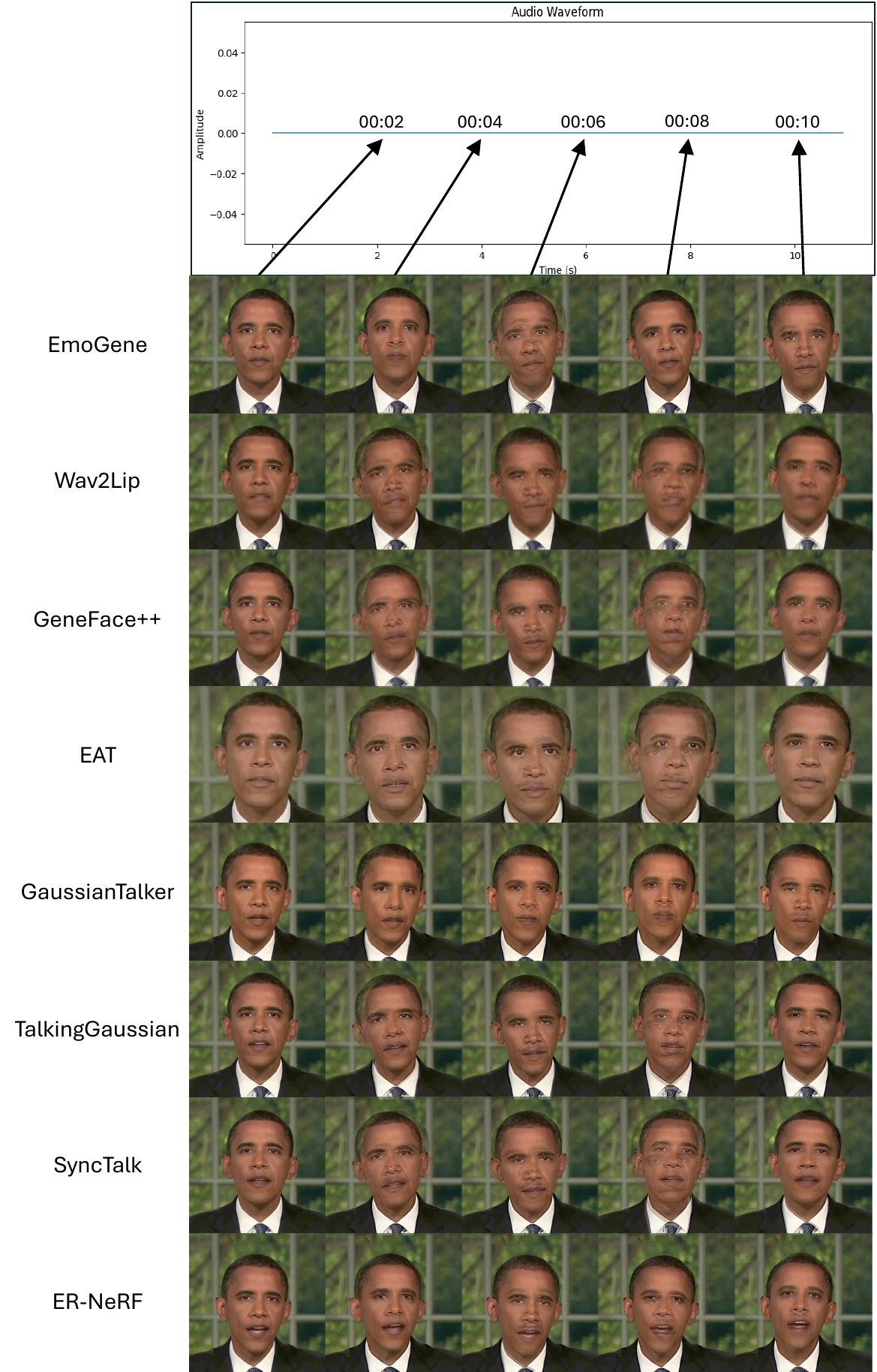}
  \caption{\textbf{Qualitative comparison of idle-state video generation.} For the given silent driving audio, idle motion results from the evaluated method are visualized by overlaying each generated keyframe with its predecessor to highlight pose changes.}
  \label{idle_audio}
\end{figure}

In addition, we show the effectiveness of our idle-pose sampling method for generating natural idle-state videos in Figure~\ref{idle_audio}. We compare keyframes generated by EmoGene and other methods, overlaying each keyframe with the preceding one to highlight pose changes. We observe that given silent audio input, the pose sampling algorithm enables EmoGene to generate smooth and consistent idle motions with subtle pose changes between keyframes. In contrast, NeRF-based methods exhibit excessive pose changes and incorrect lip synchronization. For example, GeneFace++ and SyncTalk produce apparent pose shifts between the third and fourth keyframes. ER-NeRF generates extra lip movements, leading to inaccurate lip synchronization. While 3D Gaussian Splatting-based methods produce fewer pose shifts than NeRF-based approaches, they still show noticeable changes between keyframes. Furthermore, both Wav2Lip and EAT display significant motion inconsistencies in multiple keyframes. These results demonstrate the effectiveness of our pose sampling method in producing natural and consistent idle-state videos in response to silent audio input.

\textbf{User Study.} We conducted a user study with 20 participants to systematically assess the quality of generated videos. For each of the 8 methods, we generated 3 video clips across 8 emotion categories, resulting in a total of 192 videos. We utilized the Mean Opinion Score (MOS) for evaluation, ranging from 1 (Bad) to 5 (Excellent). The participants were instructed to rate each video based on 4 criteria: 1) emotional accuracy; 2) lip synchronization; 3) video realism; and 4) video quality.

The average scores for each method are presented in Table~\ref{mos}. Our observations are as follows: 1) EmoGene outperforms the other methods in emotional accuracy, video realism, and video quality. 2) EmoGene has a lower perceived lip-synchronization accuracy compared to other methods, potentially due to its emotional landmark deformation process, which may affect lip synchronization during the generation of emotional landmarks.

\subsection{Ablation Study}
In this section, we present ablation study results to validate the necessity of the proposed components in EmoGene.

\input{tables/ablation_deform}

\input{tables/ablation_idle}

\textbf{Landmark Deformation Model.} To assess the significance of the landmark deformation model, we performed an ablation study by excluding it from the pipeline. In this setting, the framework relies solely on neutral landmarks without incorporating emotional deformation. The results are presented in Table~\ref{abla_deform}, which shows a decrease in SSIM, PSNR, emotion score, and M/F-LMD metrics. This performance drop illustrates the important role of the landmark deformation model in generating realistic emotional expression within the framework.

\textbf{Idle-State Sampling.} We ablated 4 configurations of the idle-state sampling method: 1) fixed idle length (\( \text{I}_{\text{l}})\) and randomly sampled gap length (\( \text{G}_{\text{l}})\); 2) randomly sampled idle length (\( \text{I}_{\text{l}})\) and gap length (\( \text{G}_{\text{l}})\); 3) evenly sampled idle length (\( \text{I}_{\text{l}})\) and gap length (\( \text{G}_{\text{l}})\); 4) no idle-state sampling. The results are presented in Table~\ref{abla_idle}, which indicates that EmoGene generates the most natural and consistent idle-state videos using the original idle-state sampling configuration (randomly sampled idle length with fixed gap length). This configuration produces the best scores in average motion velocity, velocity standard deviation, and average motion acceleration. These findings highlight the effectiveness of the idle-state sampling in producing smooth and coherent idle-state videos.

\section{CONCLUSION}
In this paper, we propose EmoGene for synthesizing emotion-aware talking-head videos. Our proposed landmark deformation model synthesizes robust emotional landmarks to condition the NeRF models, enabling the rendering of lifelike and vivid emotional expressions that are both emotionally accurate and visually realistic. Furthermore, a pose sampling method facilitates the generation of natural idle-state videos from silent audio, ensuring smooth and consistent motion when no speech is present. Extensive experiments demonstrate that EmoGene achieves better emotional accuracy, identity preservation, and idle-state motion naturalness.

\textbf{Limitations.} 1) The performance of EmoGene is constrained by the breadth and variety of the emotional training data. 2) The deformation of neutral landmarks to emotional landmarks may impact lip-synchronization accuracy in generated videos.

\textbf{Future Directions.}
Future work could expand the emotional training dataset to encompass a wider range of expressions to enhance generalization. To improve lip synchronization while achieving a vivid emotion generation, future works can also consider refining the landmark deformation process by incorporating more expressive generative methods. Additionally, conducting comprehensive user studies would provide deeper insights into user-perceived emotional accuracy, image fidelity, and video quality.

\section{ACKNOWLEDGMENTS}
The authors thank the generous support from bitHuman, Inc.


{\small
\bibliographystyle{ieee}
\bibliography{Paper}
}

\end{document}

%% file: tables/quant_main.tex
\begin{table*}[!t]
\centering
\caption{Quantitative results of the evaluated methods on MEAD and OOD datasets. ``↑": higher is better. ``↓": lower is better. The best results are in \textcolor{red}{red}. The second-best results are in \textcolor{blue}{blue}. Each cell shows the percentage performance difference relative to our method, EmoGene.}
\label{table_quant}
\resizebox{\linewidth}{!}{%
\begin{tabular}{@{}lcccc|cccc@{}}
\toprule
& \multicolumn{4}{c|}{MEAD} & \multicolumn{4}{c}{OOD} \\ \cmidrule(lr){2-5} \cmidrule(lr){6-9}
\multicolumn{1}{c}{Method/Score} 
    & SSIM ↑ & PSNR ↑ & \( \text{Score}_{\text{emotion}} \) (\%) ↑ & M/F-LMD ↓ 
    & SSIM ↑ & PSNR ↑ & \( \text{Score}_{\text{emotion}} \) (\%) ↑ & M/F-LMD ↓ \\ 
\midrule

EAT~\cite{eat} 
& 0.479 (-34.4\%) 
  & 16.940 (-14.0\%) 
  & \textcolor{blue}{19.389} (-1.7\%) 
  & 3.828/7.858 (-6.3\%/-32.5\%) 
& 0.387 (-27.0\%) 
  & 11.099 (-1.1\%) 
  & \textcolor{red}{19.836} (+4.1\%) 
  & \textcolor{blue}{3.823}/8.562 (-5.0\%/-17.1\%) 
\\

TalkingGaussian~\cite{talkinggaussian} 
& 0.497 (-31.9\%) 
  & 13.083 (-33.6\%) 
  & 8.479 (-57.0\%) 
  & 3.713/6.654 (-3.1\%/-12.2\%) 
& 0.516 (-2.6\%) 
  & \textcolor{red}{11.301} (+0.7\%) 
  & 12.268 (-35.6\%) 
  & 4.155/7.537 (-14.1\%/-3.1\%) 
\\

Wav2Lip~\cite{wav2lip} 
& 0.623 (-14.7\%) 
  & 19.428 (-1.4\%) 
  & 12.870 (-34.7\%) 
  & 4.043/6.318 (-12.2\%/-6.6\%) 
& 0.514 (-3.0\%) 
  & 11.227 (+0.1\%) 
  & 13.051 (-31.5\%) 
  & 4.155/7.495 (-14.1\%/-2.5\%) 
\\

ER-NeRF~\cite{ernerf} 
& 0.660 (-9.6\%) 
  & 18.275 (-7.2\%) 
  & 12.323 (-37.5\%) 
  & 4.129/7.189 (-14.6\%/-21.3\%) 
& 0.518 (-2.3\%) 
  & \textcolor{blue}{11.280} (+0.5\%) 
  & 12.430 (-34.8\%) 
  & 4.087/7.822 (-12.2\%/-7.0\%) 
\\

GeneFace++~\cite{geneface++} 
& 0.729 (-0.1\%) 
  & \textcolor{blue}{19.687} (-0.1\%) 
  & 11.834 (-40.0\%) 
  & 3.970/6.334 (-10.2\%/-6.8\%) 
& \textcolor{blue}{0.529} (-0.2\%) 
  & 11.219 (-0.0\%) 
  & 12.715 (-33.3\%) 
  & 3.908/7.327 (-7.3\%/-0.2\%) 
\\

SyncTalk~\cite{syntalk} 
& \textcolor{blue}{0.732} (+0.3\%) 
  & 19.340 (-1.8\%) 
  & 13.290 (-32.6\%) 
  & \textcolor{blue}{3.639}/6.240 (-1.0\%/-5.2\%) 
& 0.507 (-4.3\%) 
  & 11.174 (-0.4\%) 
  & 12.450 (-34.7\%) 
  & 3.843/\textcolor{blue}{7.315} (-5.6\%/-0.1\%) 
\\

GaussianTalker~\cite{gaussiantalker} 
& \textcolor{red}{0.736} (+0.8\%) 
  & 19.300 (-2.0\%) 
  & 12.750 (-35.3\%) 
  & 3.819/\textcolor{blue}{6.094} (-6.0\%/-2.8\%) 
& 0.367 (-30.8\%) 
  & 7.643 (-31.9\%) 
  & 8.440 (-56.0\%) 
  & 4.783/8.563 (-31.4\%/-17.2\%) 
\\ 
\midrule

\textbf{EmoGene (Ours) }
& 0.730 
  & \textcolor{red}{19.698} 
  & \textcolor{red}{19.715} 
  & \textcolor{red}{3.602}/\textcolor{red}{5.929} 
& \textcolor{red}{0.530} 
  & 11.220 
  & \textcolor{blue}{19.054} 
  & \textcolor{red}{3.641}/\textcolor{red}{7.309} 
\\ 
\midrule

Ground Truth 
& 1.000 
  & $\infty$ 
  & 28.297
  & 0.000/0.000
& N/A
  & N/A
  & N/A
  & N/A
\\

\bottomrule
\end{tabular}
}
\end{table*}

%% file: tables/quant_idle.tex
\begin{table}[!t]
\centering
\caption{Quantitative results of idle videos generated by the evaluated methods. The best results are in \textcolor{red}{red}. The second-best results are in \textcolor{blue}{blue}.}
\label{table_quant_idle}
\resizebox{\columnwidth}{!}{%
\begin{tabular}{@{}l|cccc@{}}
\toprule
Method/Score & Vel. Avg. ↓ & Vel. Std. ↓ & Acc. Avg. ↓ & Acc. Std. ↓ \\ \midrule
ER-NeRF~\cite{ernerf} &  0.0528 &	0.0455 &	0.0392 &	0.0313 \\
GaussianTalker~\cite{gaussiantalker} & 0.0514 &	0.0451	& 0.0417 &	0.0319 \\ 
TalkingGaussian~\cite{talkinggaussian} &  0.0451 &	0.0407 &	0.0377 &	0.0297 \\
Wav2Lip~\cite{wav2lip} &  0.0431 & 0.0399 &	0.0378 &	0.0347 \\
EAT~\cite{eat} & 0.0371 & 0.0364 & 0.031 & 0.0259 \\ 
SyncTalk~\cite{syntalk} & 0.0349 &	0.0359 &	0.0278	& 0.0257 \\  
GeneFace++~\cite{geneface++} &  \textcolor{blue}{0.030} & \textcolor{red}{0.0242} & \textcolor{blue}{0.0208} &	\textcolor{blue}{0.0153} \\ \midrule
\textbf{EmoGene (Ours)} & \textcolor{red}{0.0175} &	\textcolor{blue}{0.0305} &	\textcolor{red}{0.0068} &	\textcolor{red}{0.0129} \\ 
\bottomrule
\end{tabular}%
}
\end{table}

%% file: tables/mos.tex
\begin{table*}[!t]
\centering
\caption{User study results. The best results are in \textbf{bold}.}
\label{mos}
\resizebox{\linewidth}{!}{%
\begin{tabular}{@{}l|cccccccc@{}}
\toprule
Criteria/Method & \textbf{EmoGene (Ours)}         & EAT             & GeneFace++           & Wav2Lip            & GaussianTalker             & SyncTalk             & TalkingGaussian               & ER-Nerf         \\ 
\midrule

Emotion Acc.     
& \textbf{2.476$\pm$0.490}
& 2.357$\pm$0.482
& 2.381$\pm$0.385
& 2.339$\pm$0.380
& 2.310$\pm$0.225
& 1.935$\pm$0.190
& 2.018$\pm$0.229
& 1.327$\pm$0.131 \\

Lip Sync.        
& 2.689$\pm$0.276
& 2.946$\pm$0.191
& 2.696$\pm$0.199
& \textbf{2.946$\pm$0.184}  
& 1.994$\pm$0.147
& 1.696$\pm$0.119
& 1.732$\pm$0.135
& 1.244$\pm$0.074 \\

Video Realness  
& \textbf{2.536$\pm$0.141}  
& 2.435$\pm$0.243
& 2.417$\pm$0.154
& 2.137$\pm$0.133
& 2.042$\pm$0.233
& 1.935$\pm$0.142
& 2.018$\pm$0.125
& 1.274$\pm$0.098 \\

Video Quality   
& \textbf{2.625$\pm$0.131}  
& 2.560$\pm$0.143
& 2.583$\pm$0.146
& 2.417$\pm$0.170
& 2.452$\pm$0.076
& 2.030$\pm$0.161
& 2.185$\pm$0.090
& 1.286$\pm$0.067 \\ 
\midrule

Average
& \textbf{2.581$\pm$0.260}
& 2.575$\pm$0.265
& 2.519$\pm$0.221
& 2.460$\pm$0.217
& 2.200$\pm$0.170
& 1.899$\pm$0.153
& 1.988$\pm$0.145
& 1.283$\pm$0.093 \\

\bottomrule
\end{tabular}%
}
\end{table*}

%% file: tables/ablation_deform.tex
\begin{table}[!t]
\centering
\caption{Ablation study results of the landmark deformation model. The best results are in \textbf{bold}.}
\label{abla_deform}
\resizebox{\columnwidth}{!}{%
\begin{tabular}{@{}l|cccc@{}}
\toprule
Setting/Score  & SSIM ↑ & PSNR ↑ & \( \text{Score}_{\text{emotion}} \) (\%) ↑ & M/F-LMD ↓ \\ \midrule

w/o LDM
& 0.509
  &  10.778
  &  12.344
  &  4.440/9.476
\\ 

EmoGene
& \textbf{0.730 }
  & \textbf{19.698 }
  & \textbf{19.715}
  & \textbf{3.602}/\textbf{5.929}
\\

\bottomrule
\end{tabular}
}
\end{table}

%% file: tables/ablation_idle.tex
\begin{table}[!t]
\centering
\caption{Ablation study results on idle-state pose sampling. The best results are in \textbf{bold}.}
\label{abla_idle}
\resizebox{\columnwidth}{!}{%
\begin{tabular}{@{}l|cccc@{}}
\toprule
Setting/Score & Vel. Avg. ↓ & Vel. Std. ↓ & Acc. Avg. ↓ & Acc. Std. ↓ \\ \midrule
Fixed \( \text{I}_{\text{l}} \) \& Random \( \text{G}_{\text{l}} \) &  0.0256 & 0.0339 &	0.0084 &	0.0115 \\
Random \( \text{I}_{\text{l}} \) \& Random \( \text{G}_{\text{l}} \) &  0.0253 & 0.0338 &	0.0083 &	0.0113 \\ 
\( \text{I}_{\text{l}} \) \& \( \text{G}_{\text{l}} \) Sampled Evenly  &  0.0259 & 0.0339 &	0.0084 &	\textbf{0.0112} \\
w/o &  0.0378 &	0.0351 &	0.0125 &	0.0125 \\ \midrule
EmoGene &  \textbf{0.0175} &	\textbf{0.0305} &	\textbf{0.0068} &	0.0129 \\
\bottomrule
\end{tabular}%
}
\end{table}

%% file: Paper.bbl
\begin{thebibliography}{10}\itemsep=-1pt

\bibitem{gstalker}
B.~Chen, S.~Hu, Q.~Chen, C.~Du, R.~Yi, Y.~Qian, and X.~Chen.
\newblock Gstalker: Real-time audio-driven talking face generation via deformable gaussian splatting, 2024.

\bibitem{gaussiantalker}
K.~Cho, J.~Lee, H.~Yoon, Y.~Hong, J.~Ko, S.~Ahn, and S.~Kim.
\newblock Gaussiantalker: Real-time high-fidelity talking head synthesis with audio-driven 3d gaussian splatting, 2024.

\bibitem{voxceleb2}
J.~Chuang, A.~Nagrani, and A.~Zisserman.
\newblock Voxceleb2: Deep speaker recognition.
\newblock {\em INTERSPEECH}, 2018.

\bibitem{syncnet}
J.~Chung and A.~Zisserman.
\newblock Out of time: automated lip sync in the wild. in workshop on multi-view lip-reading.
\newblock {\em Asian Conference on Computer Vision}, pages 251--263, 2017.

\bibitem{deep3drecon}
Y.~Deng, J.~Yang, S.~Xu, D.~Chen, Y.~Jia, and X.~Tong.
\newblock Accurate 3d face reconstruction with weakly-supervised learning: From single image to image set.
\newblock {\em IEEE Computer Vision and Pattern Recognition Workshops}, 2019.

\bibitem{3dmm}
B.~Egger, W.~A. Smith, A.~Tewari, S.~Wuhrer, M.~Zollhoefer, T.~Beeler, F.~Bernard, T.~Bolkart, A.~Kortylewski, S.~Romdhani, et~al.
\newblock 3d morphable face models—past, present, and future.
\newblock {\em ACM Transactions on Graphics (ToG)}, 39(5):1--38, 2020.

\bibitem{a2l_2}
S.~E. Eskimez, R.~K. Maddox, C.~Xu, and Z.~Duan.
\newblock Generating talking face landmarks from speech, 2018.

\bibitem{wavenet}
A.~O. et~al.
\newblock Wavenet: A generative model for raw audio.
\newblock {\em arXiv:1609.03499}, 2016.

\bibitem{facial}
C.~Z. et~al.
\newblock Facial: Synthesizing dynamic talking face with implicit attribute learning.
\newblock {\em Proceedings of the IEEE/CVF International Conference on Computer Vision}, pages 3867--3876, 2021.

\bibitem{nerf2024_1}
D.~L. et~al.
\newblock Ae-nerf: Audio enhanced neural radiance field for few shot talking head synthesis.
\newblock {\em Proceedings of the AAAI Conference on Artificial Intelligence}, 38(4):3037--3045, 2024.

\bibitem{videoretalking}
K.~C. et~al.
\newblock Videoretalking: Audio-based lip synchronization for talking head video editing in the wild.
\newblock {\em SIGGRAPH-ASIA}, (30):1--9, 2022.

\bibitem{nerfo5}
K.~P. et~al.
\newblock Hypernerf: a higher-dimensional representation for topologically varying neural radiance fields.
\newblock {\em ACM Transactions on Graphics}, 40(238):1--12, 2021.

\bibitem{emo_2024_4}
K.~S. et~al.
\newblock Laughtalk: Expressive 3d talking head generation with laughter.
\newblock {\em Proceedings of the IEEE/CVF Winter Conference on Applications of Computer Vision}, pages 6404--6413, 2024.

\bibitem{mead}
K.~W. et~al.
\newblock Mead: A large-scale audio-visual dataset for emotional talking-face generation.
\newblock {\em European Conference on Computer Vision}, 12366:700--717, 2020.

\bibitem{emo_2024_2}
N.~D. et~al.
\newblock Emoportraits: Emotion-enhanced multimodal one-shot head avatars.
\newblock {\em Proceedings of the IEEE/CVF Conference on Computer Vision and Pattern Recognition}, pages 8498--8507, 2024.

\bibitem{hubert}
W.~H. et~al.
\newblock Hubert: Self-supervised speech representation learning by masked prediction of hidden units.
\newblock {\em arXiv:1312.6114}, 29:3451--3460, 2021.

\bibitem{evp}
X.~J. et~al.
\newblock Audio-driven emotional video portraits.
\newblock {\em Proceedings of the IEEE/CVF Conference on Computer Vision and Pattern Recognition}, pages 14080--14089, 2021.

\bibitem{styletalk}
Y.~M. et~al.
\newblock Speech driven talking face generation from a single image and an emotion condition.
\newblock {\em IEEE Transactions on Multimedia}, pages 3480--3490, 2021.

\bibitem{geneface++}
Z.~Y. et~al.
\newblock Geneface++: Generalized and stable real-time audio-driven 3d talking face generation.
\newblock {\em arXiv preprint arXiv:2305.00787}, 2023.

\bibitem{lip_3}
T.~Ezzat, G.~Geiger, and T.~Poggio.
\newblock Trainable videorealistic speech animation.
\newblock 21(3):388–398, July 2002.

\bibitem{eat}
Y.~Gan, Z.~Yang, X.~Yue, L.~Sun, and Y.~Yang.
\newblock Efficient emotional adaptation for audio-driven talking-head generation.
\newblock {\em Proceedings of the IEEE/CVF International Conference on Computer Vision}, pages 22634--22645, 2023.

\bibitem{adnerf}
Y.~Guo, K.~Chen, S.~Liang, Y.~Liu, H.~Bao, and J.~Zhang.
\newblock Ad-nerf: Audio driven neural radiance fields for talking head synthesis.
\newblock {\em Proceedings of the IEEE/CVF International Conference on Computer Vision}, pages 5784--5794, 2020.

\bibitem{depthaware}
F.~Hong, L.~Zhang, L.~Shen, and D.~Xu.
\newblock Depth-aware generative adversarial network for talking head video generation.
\newblock {\em Proceedings of the IEEE/CVF Conference on Computer Vision and Pattern Recognition}, pages 3397--3406, 2022.

\bibitem{a2l_1}
F.-T. Hong, Y.~Liu, Y.~Li, C.~Zhou, F.~Yu, and D.~Xu.
\newblock Dreamhead: Learning spatial-temporal correspondence via hierarchical diffusion for audio-driven talking head synthesis, 2024.

\bibitem{eamm}
X.~Ji, H.~Zhou, K.~Wang, Q.~Wu, W.~Wu, F.~Xu, and X.~Cao.
\newblock Eamm: One-shot emotional talking face via audio-based emotion-aware motion model.
\newblock {\em ACM SIGGRAPH 2022 Conference Proceedings}, (61):1--10, 2022.

\bibitem{3dgs}
B.~Kerbl, G.~Kopanas, T.~Leimkühler, and G.~Drettakis.
\newblock 3d gaussian splatting for real-time radiance field rendering, 2023.

\bibitem{vae}
D.~Kingma and M.~Welling.
\newblock Auto-encoding variational bayes.
\newblock {\em arXiv:1312.6114}, 2013.

\bibitem{talkinggaussian}
J.~Li, J.~Zhang, X.~Bai, J.~Zheng, X.~Ning, J.~Zhou, and L.~Gu.
\newblock Talkinggaussian: Structure-persistent 3d talking head synthesis via gaussian splatting, 2024.

\bibitem{ernerf}
J.~Li, J.~Zhang, X.~Bai, J.~Zhou, and L.~Gu.
\newblock Efficient region-aware neural radiance fields for high-fidelity talking portrait synthesis.
\newblock In {\em Proceedings of the IEEE/CVF International Conference on Computer Vision (ICCV)}, pages 7568--7578, October 2023.

\bibitem{emoface}
C.~Liu, Q.~Lin, Z.~Zeng, and Y.~Pan.
\newblock Emoface: Audio-driven emotional 3d face animation.
\newblock In {\em 2024 IEEE Conference Virtual Reality and 3D User Interfaces (VR)}, page 387–397. IEEE, Mar. 2024.

\bibitem{livespeechportraits}
Y.~Lu, J.~Chai, and X.~Cao.
\newblock Live speech portraits: Real-time photorealistic talking-head animation.
\newblock {\em ACM Transactions on Graphics}, 40(220):1--17, 2021.

\bibitem{nerf}
B.~Mildenhall, P.~Srinivasan, M.~Tancik, J.~Barron, R.~Ramamoorthi, and R.~Ng.
\newblock Nerf: Representing scenes as neural radiance fields for view synthesis.
\newblock {\em Proceedings of the IEEE/CVF International Conference on Computer Vision}, pages 405--421, 2020.

\bibitem{nocentini2023learninglandmarksmotionspeech}
F.~Nocentini, C.~Ferrari, and S.~Berretti.
\newblock Learning landmarks motion from speech for speaker-agnostic 3d talking heads generation, 2023.

\bibitem{syntalk}
Z.~Peng, W.~Hu, Y.~Shi, X.~Zhu, X.~Zhang, H.~Zhao, J.~He, H.~Liu, and Z.~Fan.
\newblock Synctalk: The devil is in the synchronization for talking head synthesis, 2024.

\bibitem{emotalk}
Z.~Peng, H.~Wu, Z.~Song, H.~Xu, X.~Zhu, J.~He, H.~Liu, and Z.~Fan.
\newblock Emotalk: Speech-driven emotional disentanglement for 3d face animation, 2023.

\bibitem{wav2lip}
K.~Prajwal, R.~Mukhopadhyay, V.~Namboodiri, and C.~Jawahar.
\newblock A lip sync expert is all you need for speech to lip generation in the wild.
\newblock {\em Proceedings of the 28th ACM International Conference on Multimedia}, pages 484--492, 2020.

\bibitem{a2l_3}
Z.~Qi, X.~Zhang, N.~Cheng, J.~Xiao, and J.~Wang.
\newblock Difftalker: Co-driven audio-image diffusion for talking faces via intermediate landmarks, 2023.

\bibitem{detectoutofdistri}
X.~Ran, M.~Xu, L.~Mei, Q.~Xu, and Q.~Liu.
\newblock Detecting out-of-distribution samples via variational auto-encoder with reliable uncertainty estimation.
\newblock {\em Neural Networks}, 145:199--208, 2022.

\bibitem{nerf2024_2}
A.~Shin, J.~Lee, J.~Hwang, Y.~Kim, and G.~Park.
\newblock Wav2nerf: Audio-driven realistic talking head generation via wavelet-based nerf.
\newblock {\em Image and Vision Computing}, 148, 2024.

\bibitem{emo_2024_5}
Z.~Sun, Y.~Xuan, F.~Liu, and Y.~Xiang.
\newblock Fg-emotalk: Talking head video generation with fine-grained controllable facial expressions.
\newblock {\em Proceedings of the AAAI Conference on Artificial Intelligence}, 38(5):5043--5051, 2024.

\bibitem{taigman2014deepface}
Y.~Taigman, M.~Yang, M.~Ranzato, and L.~Wolf.
\newblock Deepface: Closing the gap to human-level performance in face verification.
\newblock In {\em Proceedings of the IEEE conference on computer vision and pattern recognition}, pages 1701--1708, 2014.

\bibitem{emo_2024_3}
S.~Tan, B.~Ji, and Y.~Pan.
\newblock Flowvqtalker: High-quality emotional talking face generation through normalizing flow and quantization.
\newblock {\em Proceedings of the IEEE/CVF Conference on Computer Vision and Pattern Recognition}, pages 26317--26327, 2024.

\bibitem{lip_2}
K.~Vougioukas, S.~Petridis, and M.~Pantic.
\newblock Realistic speech-driven facial animation with gans, 2019.

\bibitem{talklip}
J.~Wang, X.~Qian, M.~Zhang, R.~T. Tan, and H.~Li.
\newblock Seeing what you said: Talking face generation guided by a lip reading expert, 2023.

\bibitem{oneshot}
S.~Wang, L.~Li, Y.~Ding, and X.~Yu.
\newblock One-shot talking face generation from single-speaker audio-visual correlation learning.
\newblock {\em Proceedings of the AAAI Conference on Artificial Intelligence}, 36(3):2531--2539, 2020.

\bibitem{lip_1}
O.~Wiles, A.~S. Koepke, and A.~Zisserman.
\newblock X2face: A network for controlling face generation by using images, audio, and pose codes, 2018.

\bibitem{pointtalk}
Y.~Xie, T.~Feng, X.~Zhang, X.~Luo, Z.~Guo, W.~Yu, H.~Chang, F.~Ma, and F.~R. Yu.
\newblock Pointtalk: Audio-driven dynamic lip point cloud for 3d gaussian-based talking head synthesis, 2024.

\bibitem{geneface}
Z.~Ye, Z.~Jiang, Y.~Ren, J.~Liu, J.~He, and Z.~Zhao.
\newblock Geneface: Generalized and high-fidelity audio-driven 3d talking face synthesis.
\newblock {\em arXiv preprint arXiv:2301.13430}, 2023.

\bibitem{a2l_4}
S.~Zhai, M.~Liu, Y.~Li, Z.~Gao, L.~Zhu, and L.~Nie.
\newblock Talking face generation with audio-deduced emotional landmarks.
\newblock {\em IEEE Transactions on Neural Networks and Learning Systems}, 35(10):14099--14111, 2024.

\bibitem{flowguided}
Z.~Zhang, L.~Li, Y.~Ding, and C.~Fan.
\newblock Flow-guided one-shot talking face generation with a high-resolution audio-visual dataset.
\newblock {\em Proceedings of the IEEE/CVF Conference on Computer Vision and Pattern Recognition}, pages 3661--3670, 2021.

\bibitem{tk_1}
H.~Zhou, Y.~Liu, Z.~Liu, P.~Luo, and X.~Wang.
\newblock Talking face generation by adversarially disentangled audio-visual representation.
\newblock AAAI'19/IAAI'19/EAAI'19. AAAI Press, 2019.

\bibitem{pc_avs}
H.~Zhou, Y.~Sun, W.~Wu, C.~Loy, X.~Wang, and Z.~Liu.
\newblock Pose-controllable talking face generation by implicitly modularized audio-visual representation.
\newblock {\em Proceedings of the IEEE/CVF Conference on Computer Vision and Pattern Recognition}, pages 4174--4184, 2021.

\bibitem{makeittalk}
Y.~Zhou, X.~Han, E.~Shechtman, J.~Echevarria, E.~Kalogerakis, and D.~Li.
\newblock Makeittalk: Speaker-aware talking-head animation.
\newblock {\em ACM Transactions on Graphics}, (221):1--15, 2020.

\end{thebibliography}
